\documentclass[a4paper,11pt,twocolumn,fleqn]{article}

\usepackage[margin=2.5cm,heightrounded=true]{geometry}
\usepackage[ngerman,english]{babel}

\usepackage{tgtermes}

\usepackage[fleqn]{amsmath}
\usepackage{newtxmath}
\usepackage{inconsolata}
\usepackage[utf8]{inputenc}
\usepackage[TS1,T1]{fontenc}
\def\longs{{\fontencoding{TS1}\selectfont s}}
\usepackage{xcolor}

\usepackage[pdfencoding=auto,psdextra,bookmarksopen]{hyperref}
\usepackage{xcolor}
\definecolor{darkblue}{rgb}{0, 0, 0.5}
\hypersetup{colorlinks=true, citecolor=darkblue, linkcolor=darkblue, urlcolor=darkblue}
\usepackage{graphicx}
\usepackage{authoraftertitle}
\usepackage{url}
\usepackage[nobiblatex]{xurl}
\usepackage{longtable}
\usepackage{booktabs}
\usepackage{multirow}
\usepackage{siunitx}
\DeclareSIUnit{\nothing}{\relax}
\DeclareMathOperator*{\argmax}{arg\,max}
\usepackage{mathtools}
\usepackage[textsize=tiny]{todonotes}
\usepackage{subcaption}
\usepackage[raggedright]{titlesec}

\renewenvironment{abstract}%
  {\begin{center}\large\textbf{\abstractname}\end{center}%
    \begin{list}{}%
      {\setlength{\rightmargin}{0.6cm}%
        \setlength{\leftmargin}{0.6cm}}%
      \item[]\ignorespaces%
      \small
  }%
  {\unskip\end{list}}

\usepackage[style=authoryear,backend=biber,mergedate=minimum,uniquelist=false,maxcitenames=2,maxbibnames=99,dashed=false,useprefix=true,urldate=long,doi=false,url=false]{biblatex}
\input{biblatex_macros.def}

\usepackage[german=guillemets,autostyle=true]{csquotes}
\SetCiteCommand{\autocite}
\usepackage[format=plain,labelfont=bf,justification=raggedright]{caption}
\usepackage[inline]{enumitem}
\setlist{itemsep=0pt,parsep=4pt}
\setlist[enumerate]{label=\arabic*., itemsep=0pt}

\newcommand{\joinchar}{\textcolor{lightgray}{\rule[-1pt]{8pt}{10pt}}}
\newcommand{\splitchar}{\rule[-1.5pt]{6pt}{1.5pt}}

\newcommand{\graph}[1]{$\langle$#1$\rangle$}

\author{\bf Anton Ehrmanntraut\\[1ex]Julius-Maximilians-Universität Würzburg\\\small\texttt{anton.ehrmanntraut@uni-wuerzburg.de}}
\date{\today}
\title{\bf Historical German Text Normalization Using Type- and Token-Based Language Modeling}

\begin{document}

\maketitle

\begin{abstract}
Historic variations of spelling poses a challenge for full-text search or natural language processing on historical digitized texts. To minimize the gap between the historic orthography and contemporary spelling, usually an automatic orthographic normalization of the historical source material is pursued. This report proposes a normalization system for German literary texts from c.~1700–1900, trained on a parallel corpus. The proposed system makes use of a machine learning approach using Transformer language models, combining an encoder-decoder model to normalize individual word types, and a pre-trained causal language model to adjust these normalizations within their context. An extensive evaluation shows that the proposed system provides state-of-the-art accuracy, comparable with a much larger fully end-to-end sentence-based normalization system, fine-tuning a pre-trained Transformer large language model. 
However, the normalization of historical text remains a challenge due to difficulties for models to generalize, and the lack of extensive high-quality parallel data.
\end{abstract}

\section{Introduction}

Ongoing digitization efforts are making more and more historical documents digitally available, such as historical newspaper, journals, treatises, novels, and so on. These digital editions do not only consist of the scan, but also present the text in a machine-readable form, as generated by automatic optical character recognition (OCR).
In principle, this allows researchers to, e.g., 1) perform a full-text search in their digitized corpora, or 2) to perform \emph{distant reading}: using Natural Language Processing (NLP) tools, automatically process documents in order to make analyses over large corpora.

One major obstacle to this research design is language change. The older the historical texts are, the more they diachronically deviate from the current standard language in orthography, syntax, lexicon, etc. Additionally, we see synchronic variability, in particular on orthography:  prior to a standardization of orthography (which in itself is a fairly recent phenomenon; for instance, this happened for the German language in 1901) the range of spelling variations was considerably wider.
Already these changes on the level of orthography, diachronically and synchronically, pose a problem for the above-mentioned downstream operations. First, a complete full-text search would require users to know and specify all variations of the search term they are interested in.
For instance, in documents of the \emph{Deutsche Textarchivs}\footnote{\url{https://deutschestextarchiv.de/}} (DTA) between 1830 and 1890, we can observe the words \emph{heirathen, heiraten, heurathen, heyrathen}, all variants of the synchronically active extant form \emph{heiraten} which would need to be specified by the user in order to be retrieved.
Second, the use of NLP gets significantly hindered since off-the-shelf NLP tools are primarily trained on contemporary language data, therefore downstream accuracy drops when applied to out-of-domain historical data, like depicted in Fig.~\ref{fig:originalprint}.

\begin{figure*}[t]
    \strut\hfill\begin{subfigure}[t]{0.45\textwidth}
        \raggedright\em
    Das Bedeutend\longs{}te jedoch was die Freunde mit \longs{}tiller Aufmerk\longs{}amkeit beobachteten, war, \underline{daß} Ottilie den \underline{Coffer} zum \underline{er\longs{}tenmal} ausgepackt und daraus \underline{ver\longs{}chiedenes} gew$\smash{\overset{\text{\tiny\kern2pt e}}{\text{a}}}$hlt und abge\longs{}chnitten hatte, was zu einem einzigen aber ganzen und vollen Anzug hinreichte. Als \longs{}ie das \underline{Uebrige} mit \underline{Beyh$\overset{\text{\tiny\kern2pt e}}{\text{u}}$lfe} Nannys wieder einpacken wollte, konnte \longs{}ie kaum damit \underline{zu Stande kommen}; der Raum war $\overset{\text{\tiny\kern2pt e}}{\text{u}}$ebervoll, obgleich \longs{}chon ein \underline{Theil} herausgenommen war. 
\vspace*{.2cm}
\caption{\raggedright Diplomatic transcription of the first print of \emph{Die
Wahlverwandtschaften} (1831).}\label{fig:originalprint}
    \end{subfigure}\hfill
    \begin{subfigure}[t]{.45\textwidth}\raggedright\em
        Das Bedeutendste jedoch, was die Freunde mit stiller Aufmerksamkeit beobachteten, war, \underline{dass} Ottilie den \underline{Koffer} zum \underline{ersten Mal} ausgepackt und daraus \underline{Verschiedenes} gewählt und abgeschnitten hatte, was zu einem einzigen aber ganzen und vollen Anzug hinreichte. Als sie das \underline{Übrige} mit \underline{Beihilfe} Nannys wieder einpacken wollte, konnte sie kaum damit \underline{zustandekommen}; der Raum war  übervoll,\linebreak obgleich schon ein \underline{Teil} herausgenommen\linebreak war. 
\vspace*{.2cm}
        \caption{\raggedright Normalization to contemporary orthography.}\label{fig:normalprint}
    \end{subfigure}\hfill\strut\\
\caption[]{Comparison between a OCR'd historical print and a potential example of its normalized equivalent. Underlined spans correspond to changes (trivial transliterations of long-s and superscript-e are ignored).\footnotemark}\label{fig:parallel}
\end{figure*}

One possible solution to this problems is (historical) \emph{text normalization} (also known as \emph{canonicalization}, \emph{modernization}; \cite{piotrowski_natural_2012}): automatically “translate” the historical source text into a contemporary orthography by replacing each word in its historical spellings with its respective extant canonical form. This 1) removes spelling variation within the documents, and 2) reduces the linguistic gap between the historical state of the language and the contemporary state. In the example above, for instance, this normalization would replace all variants \emph{heyrathen}, \emph{heiraten}, \ldots{} with the canonical \emph{heiraten}, thus a single search term would retrieve all relevant occurrences. In the same way, this normalization process would modify the spelling in Fig.~\ref{fig:normalprint} to conform to current standards, making it more compatible with NLP tools. We have empirical evidence that such normalization indeed helps when performing downstream tasks like Part-of-Speech Tagging, Dependency Parsing, Named Entity Recognition, or Sentiment Analysis \parencites{sang_clin27_2017}{van_der_goot_multilexnorm_2021}{kogkitsidou_normalisation_2020}{bucur_sequence--sequence_2021}.

In fact, such normalization practice is quite common. In the “analog world” of editions, publishers and editors frequently normalize orthography and punctuation by hand in their contemporary editions of historical works (although with the motivation to ease reading for their intended audience). In the digital NLP domain, this text normalization is also an established topic, aiming to perform this task automatically: previous work did not only focus on \emph{historical} text normalization like discussed here \parencites{jurish_finite-state_2012}{bollmann_normalization_2018}{bollmann_large-scale_2019}{bawden_automatic_2022}, but also on normalization of computer-mediated communication \parencites{han_lexical_2011}{van_der_goot_multilexnorm_2021} such as in instant messages or tweets 
\begin{quote}
\linespread{1.0}\selectfont
    \emph{wollte ich zu mindestens net und ich geh jetzt meine schuhe hole bis später hdggggdl} 
\end{quote}
and on normalization of dialects to their respective standard varieties (\cites{scherrer_automatic_2016}{kuparinen_dialect--standard_2023}):
\begin{quote}
\linespread{1.0}\selectfont
        \emph{ich wäiss aber nüme wele leerer das gsii isch}. 
\end{quote}
\footnotetext{Facsimile is available at \url{https://www.deutschestextarchiv.de/book/view/goethe_wahlverw02_1809?p=316}.}

This work will focus on the historical text normalization of German literary texts from c.~1700 up to 1900. 
For one, because Computational Literary Studies (CLS; cf.~\cites{jannidis_einführung_2022}) and Computational Stylistics \parencite[cf.][]{herrmann_computational_2021} are hot subfields of Digital Humanities, both employing NLP tools for analyses \parencites{hatzel_machine_2023}{schoch_clsinfra_2023} and therefore interested in automatic normalization methods.
For another, unlike other historical text domains such as newspapers, there is a tradition of normalization in literary editions. Like in Fig.~\ref{fig:parallel}, editors manually normalized the text of historical prints to contemporary standards for their editions. Along with the OCR'd original historical prints, this allows us to construct parallel corpora between historical and normalized forms, which can be used as a training dataset to learn this (implicit) normalization.

Although there is a strong need in CLS for text normalization systems pre-processing their literary datasets for downstream NLP tasks, there is no easy-to-use method available. For the register considered here—German historical texts starting from c.~1700—the de-facto standard system is the \emph{Cascaded Analysis Broker} (CAB; \cite{jurish_finite-state_2012}), a closed-source system based on finite state automata, from a ruleset that has been manually designed by a linguistic expert, and provided by the DTA, available though a web interface and an API.\footnote{\url{https://deutschestextarchiv.de/public/cab/}} Several projects use CAB to normalize their historic texts, and the DTA itself uses CAB to normalize their OCR'd scans to make them searchable in their full-text search.
There are several drawbacks with this tool: 1) CAB is closed-source and cannot be adapted to one's specific need or source corpora. This also means that 2) CAB cannot be run locally, hence users are always dependent on the DTA to keep the system available. Furthermore, 3) although CAB is in frequent use, it has not been thoroughly evaluated before.\footnote{There is an evaluation of CAB by \textcite{jurish_more_2010}, but this did not measure the system's accuracy on the token level; instead his experiment assumed a downstream information retrieval scenario.}\looseness=-1

Instead of hand-written rulesets, contemporary methods to text normalization follow the now usual NLP paradigm of machine learning and Transformers. 
In comparison to expert systems, this comes with advantages since, for instance, machine learning should in principle be able to pick up unknown patterns from large training data and generalize better to unknown words. In this setting, the actual approaches to normalization can be broadly subdivided into two sides: 1) \emph{sentence-based} approaches understand text normalization like any other text translation task and utilize Transformer-based Large Language Models (LLMs) in a sequence-to-sequence fashion. The alternative 2) approaches are \emph{type-based}. These make use of the fact that in many text normalization tasks—including the setting presented here—the word order remains stable. Therefore, these approaches focus on the individual words, or more precise, types, as basic unit, and are normalizing word per word. 

The drawback to this approach, however, is its inability to handle ambiguous normalizations. This means that tokens of the same type may be normalized differently depending on the context. For example, the historic word form \emph{wehrt} appears twice in the DTA and needs to be normalized differently in each instance:
\begin{enumerate}[label=(\arabic*)]
\linespread{1.0}\selectfont
\item \emph{Wenn Ihre ganze Garderobe \longs{}o viel \underline{wehrt} i\longs{}t, als mein einziger Rock, [...]}\footnote{Ludwig Börne, \emph{Gesammelte Schriften. Zwölfter Teil} (Offenbach: Brunett, 1933), 245. Facsimile: \url{https://www.deutschestextarchiv.de/book/view/boerne_paris04_1833/?hl=wehrt&p=259}} 
        (“If your wardrobe is as valuable as my skirt, then [...]”)
\item \emph{Die junge Braut \underline{wehrt} \longs{}ich, Alles anzunehmen, was die künftige Schwiegermutter ihr geben will.}\footnote{Ludwig Otto, \emph{Zwischen Himmel und Erde} (Frankfurt am Main: Meidinger), 319. Facsimile: \url{https://www.deutschestextarchiv.de/book/view/ludwig_himmel_1856/?hl=wehrt&p=328}} 
        (“The young bride resists accepting  anything the future mother-in-law wants to give her.”)
\end{enumerate}
In the first example, \emph{wehrt} needs to be normalized to contemporary adjective \emph{wert} (“valuable”), whereas in the second example, should be normalized to contemporary verb \emph{wehrt} (“resists”). The correct normalization can only be determined by the surrounding context.

To address these kinds of ambiguous normalizations, type-based approaches are typically extended to a hybrid model, where the type normalization is followed by a second processing step, where the normalization hypotheses of each type are combined with the source text to finally derive a target text in a context-aware manner. 

\paragraph{Contribution.}

Concerning the literary domain presented here, an off-the-shelf usable Transformer-based system for historical text normalization is missing.
Therefore, the primary motivation for this work is to provide an improvement over the currently used normalization system CAB, by developing such a Transformer-based system, designed for literary language from c.~1700–1900.
In fact, \textcite{bracke} is working on a normalization system \emph{Transnormer} for a similar domain (German language starting from c.~1600), fine-tuning a pre-trained LLM, modeling the text normalization task as a translation task and thus following the sentence-based approach.

There is no general consensus on whether sentence-based or type-based systems are better–both come with their strength and weaknesses: Type-based models cannot (trivially) make use of the context to handle ambiguous normalization cases, whereas sentence-based models are typically larger and tend to require more training examples, since in natural language, almost all types have very low frequency. In contrast, trained on types (independent of their frequency), type-based normalizers utilize the linguistic diversity present in training material in a far more efficient manner.

As such, it is an open hypothesis whether type-based systems might be comparable to sentence-based Transformers, or even outperforming them. This is the research hypothesis of the report: considering the task of historical text normalization in the literary domain, how do different architectures perform? In particular, can type-based models outperform sentence-based ones, and how much do they benefit when extended to a \emph{hybrid} setup in order to resolve ambiguities?

As such, this work designs a hybrid type-based model: in a first stage, all historical types are collected and individually normalized by a encoder-decoder Transformer in isolation, ignoring the context; this normalization generates not one but multiple normalization hypotheses for each type. Then, in a second stage, the system normalizes sentence by sentence, using a causal language model that selects the appropriate normalization hypotheses for every token, ranking them on  the likelihood of the normalized sentences induced by the selection, thus allowing the the system to handle ambiguous cases.

The model is trained on the existing token-aligned parallel corpus \emph{DTA EvalCorpus}\footnote{\url{https://kaskade.dwds.de/~moocow/software/dtaec/}} \parencite{jurish_constructing_2013} between historical literary documents and contemporary editions of these documents that were manually normalized by the editors.

Thus, this work contributes the following:
\begin{enumerate}[midpenalty=0]
    \item It provides a general-purpose system for historical text normalization with state-of-the-art performance, that can be used out of the box, is publicly available and open-source, can be run locally, and even within a CPU-only setup.\footnote{Code: \url{https://github.com/aehrm/hybrid_textnorm}; Model:\url{https://huggingface.co/aehrm/dtaec-type-normalizer}.}
    \item The work provides a large-scale quantitative evaluation of the accuracy of the proposed hybrid system and several other baseline systems, including CAB and Transnormer, on a large literary dataset. The evaluation also indicates that current state-of-the-art models are still struggling with this task, still making about 1 error per 100 words. Historical text normalization is not “solved”.
    \item This evaluation also includes a comparison between Transformer-based normalizers, based on sentence-level, type-level and the proposed hybrid one, providing evidence that hybrid systems improve upon purely type-based model, and that they provide an efficient way to perform text normalization while reaching state-of-the-art performance.
\end{enumerate}


\paragraph{Outline.}

After this introduction, the following Section~\ref{sec:definition} specifies the task of historical text normalization in detail and discusses its specifies, in particular arising from the definition of the task induced by the parallel corpora.
Then, in Section~\ref{sec:related} we will review existing techniques of historical text normalizations, focusing on the different type-based and sentence-based methods.
In preparation to the model, Section~\ref{sec:dataset} explains the dataset construction from the DTA EvalCorpus, and the full model specification is presented in Section~\ref{sec:method}.
Then, in Section~\ref{sec:evaluation}, the presented model will be evaluated against several other baselines, including CAB presented above, with respect to the parallel dataset.
The work concludes with a summary and arising open questions in Section~\ref{sec:conclusion}.

\section{Problem Statement}\label{sec:definition}

Intuitively, the problem of historical text normalization as discussed in this work can be stated as follows: given a sentence in its historic form, translate the sentence into its orthographically “standard” form.
As usual, text normalization is on the orthographic level only, maintaining the lexical roots and morphosyntactic features; it does \emph{not} consist of changing word order, adding, or removing words. \parencites(cf.)(){van_der_goot_multilexnorm_2021}{jurish_finite-state_2012}

Of course, this would require us to specify what “standard” means.
For pragmatic reasons, this work avoids a clear specification, but instead follows the usual supervised machine learning setting and \emph{extensionally} defines “standard” over a parallel corpus, which consists of a set of sentences in historic form and their corresponding normalized forms given by human annotators.
We treat these normalized form as gold data and define the task by the content of the dataset, only reproducing the gold annotations of the test dataset through learning on the examples of the training dataset. The underlying guidelines are not of interest for us, and we also do not need to explicitly define what we mean by “standard”. 
This work will use the parallel corpus DTA EvalCorpus: on the source side, we have historical literary documents; on the target side, we have contemporary editions of these documents, where the editors provide a orthographic normalization of the source documents. (A detailed description of the corpus is provided in the following Sec.~\ref{sec:dataset}.) As such, following the extensional definition, the task at hand consists of reproducing the linguistic intuition of the editors that have developed the contemporary editions.

It should be highlighted that is not the only way to define text normalization: in fact, some parallel corpus projects (e.g., \cite{dipper_anselm_2013} for Middle High German, \cite{samardzic-etal-2016-archimob} for Swiss dialects)  make their normalization guidelines more explicit. Theoretically, we could build a system just using the guidelines as \emph{intensional} definition of the text normalization task, trying to reproduce the annotation process of human annotators, given the guidelines. This could be in the form of a rule-based system, as in the case of CAB. Or, in principle, we could also give the guidelines as prompt to a very large language model with general reasoning abilities, say, GPT4, to perform text normalization in a zero-shot fashion. At least in the case of this work, however, this is not possible since we do not have such guidelines available. In fact, the editions that were the basis for the target side of the parallel corpus frequently do not make their normalization guidelines explicit.

Nevertheless, even without explicit guidelines, we still can observe some characteristics of the text normalization presented here, both theoretically in terms of the general use case and register, and specifically present in the considered corpus.  The aim of this discussion is not to uncover some (unwritten) normalization guideline, but rather to place this work into the broader context of the field of text normalization. Between different datasets, use cases, and registers, the extent of the “normalization work” varies, and as such, making automatic normalization more or less difficult. In order to evaluate different normalization methods and to, e.g, be able to assess their applicability to other domains, we need to be aware of what kind of normalization this particular task entails, which linguistic dimensions are affected, but also which kinds of normalizations are excluded \parencite[cf.][19]{bollmann_normalization_2018}.

While language change generally affects all aspects of language—orthography, morphology, morphosyntax, syntax, lexicon, semantics, etc.—we observe that the literary historical texts being considered are relatively recent in terms of language change, and that all of these areas have largely stabilized in the German language, compared to contemporary language, the major exception being orthography and lexicon. As is already established by the definition of the task, it can be assumed that during normalization, the lexical roots and morphosyntactic features will remain intact, and normalization will only need to be applied at the level of spelling, like in the above example of \emph{heiraten} and its historic variants.

As such, text normalization as discussed here generally prefers maintaining lexemes even if they are considered archaic, like \emph{turbieren}, \emph{Diwan}, \emph{honett}; thus the normalization task does not necessarily require vast lexical knowledge.
Additionally on the syntactic level, we observe a tendency to keep archaic constructions for the benefit of maintaining word order, e.g. subordinate clauses without finite verb (\emph{afinite Nebensatzkonstruktion}, \emph{“Er ist so schnell gegangen, wie er gekommen.”}). Again, this means that the normalization task does not require (much) linguistic capabilities on the level of syntax.

Besides changes in individual characters, a large portion of the “normalization work” on the dataset considered here affects compound writing: \emph{statt zu finden} vs.\@ \emph{stattzufinden}, \emph{er\longs{}tenmal} vs.\@ \emph{ersten Mal} etc. In fact, this is the area where we face ambiguous cases the most frequent, since these are typically dependent on context and part-of-speech. Thus, relatively much linguistic knowledge and contextual information is needed to disambiguate these cases, for instance in the case “\emph{Lass uns den Brief zusammenschreiben}” (“let us compile the letter”; \emph{zusammenschreiben} as verb in idiomatic use) vs.\@  “\emph{Lass uns den Brief zusammen schreiben}” (“let us write the letter together”; \emph{zusammen} as adverb), or “\emph{Sie begann, laut aufzulachen}” (“she started to laugh out loudly”; \emph{aufzulachen} as verb) but “\emph{Sie hört auf zu lachen}” (“she stopped laughing”; \emph{auf zu lachen} as preposition, particle, verb). This is further complicated by the German spelling reform of 1997, which prescribes many prior  linguistic practices of writing groups of words as single compound as “incorrect”, and tends to prefer separate writing, especially when individual elements of the word group have different grammatical functions (cf.\@ \emph{soviel steht fest} vs.\@ \emph{so viel steht fest}).

Concerning the corpus, we should already remark at this point that the parallel corpus is constructed from multiple different editions, all of which approach text normalization slightly different. As such, the corpus—resp.\@ the normalization task induced by the corpus—has no uniform standard on how radically or conservatively historic forms are normalized. This concerns phenomena such as dativ-e (\emph{im Walde} vs.\@ \emph{im Wald}), but also the restoration of elisons (\emph{ew’gen} vs.\@ \emph{ewigen}), foreign words (\emph{Sauce} vs.\@ \emph{Soße}, \emph{Crucifix} vs.\@ \emph{Kruzifix}, \emph{Ballettänzerin} vs.\@ \emph{Balletttänzerin}, \emph{Lieutenant} vs.\@ \emph{Leutnant}), or named entities (\emph{Cölln}, \emph{Brandebourg}, \emph{Meyer}).

Furthermore, in the literary domain, there are several areas of text normalization for which it is unclear how to proceed theoretically, and likely disagreement between the different editors on how to approach these. In particular, how are nonstandard varieties in direct speech handled? Sometimes, these may be meaningful from a literary perspective, e.g., hinting at the speaking character's regional origin, class, or age. Or, how is poetic writing handled, which deliberately removes or adds sounds to word in order to fit a desired metric?
Concerning the first question, in this work we will ignore this kind of questions, maintaining the pragmatic position that the editions' normalization is the “gold” standard. Concerning the second question, we focus on non-poetic writing and any works of poetry will be excluded from the corpus (see Sec.~\ref{sec:dataset}).

After discussing the specific characteristics of text normalization, a formal definition of the task's structure can now be given:

\begin{quote}\em
    Transform a historical sentence, given as a sequence of $n$ tokens $w_1w_2\cdots w_n$, to a normalized sentence, also consisting of $n$ tokens $\tilde{w}_1\tilde{w}_2\cdots \tilde{w}_n$. 
\end{quote}
Based on the aforementioned characteristics, it can be generally observed that for each pair $(w_i, \tilde{w}_i)$, converting from the historical $w_i$ to the normalized $\tilde{w}_i$ only requires minor orthographic modifications. This means that only 1:1 mappings are possible. To account for normalizations in spacing, an encoding technique by \textcite{bawden_automatic_2022} is adopted, which introduces two special pseudo-characters \texttt{\splitchar{}} and \texttt{\joinchar{}}, to be used in the \emph{target} side.\footnote{U+2581 Lower one-eighth Block, U+2591 Light Shade.} The \texttt{\splitchar{}} in infix position of a token $\tilde{w}_i$ indicates that $\tilde{w}_i$ should be split at that position and is rendered as whitespace. The \texttt{\joinchar{}} in suffix position of a token $\tilde{w}_i$ indicates that the token should be joined with the following token $\tilde{w}_{i+1}$. An example of this can be seen in Fig.~\ref{fig:pseudo-chars}. In this context, an automatic normalization system would need to translate \texttt{erstenmal} to \texttt{ersten\splitchar{}mal}, and \texttt{irgend} to \texttt{irgend\joinchar{}}. By using this encoding scheme, the simplified 1:1 input/output form can be maintained, making it particularly easy to implement spacing normalization for models that only take an individual token as input.

\begin{figure*}[t]\centering
\setlength{\tabcolsep}{0pt}
\small
\begin{tabular}{l@{\hspace{10pt}}lllllllllll}
        \bfseries Src sentence & & \multicolumn{10}{l}{\emph{Zum erstenmal nahm er irgend ein Gefühl wahr.}}\\
        \bfseries Tgt sentence & & \multicolumn{10}{l}{\emph{Zum ersten mal nahm er irgendein Gefühl wahr.}}\\
        \bfseries Src tokens & \texttt{[} & \texttt{"Zum",} & \texttt{"erstenmal",} & \texttt{"nahm",} & \texttt{"er",} & \texttt{"irgend",} & \texttt{"ein",} & \texttt{"Gefühl",} & \texttt{"wahr",} & \texttt{"."} &\texttt{]}\\
        \bfseries Tgt tokens & \texttt{[} & \texttt{"Zum",} & \texttt{"ersten\splitchar{}mal",} & \texttt{"nahm",} & \texttt{"er",} & \texttt{"irgend\joinchar{}",} & \texttt{"ein",} & \texttt{"Gefühl",} & \texttt{"wahr",} & \texttt{"."} &\texttt{]}\\
\end{tabular}\vspace*{-.1cm}
    \caption{Example usage of the spacing encoding scheme using the pseudo-characters \texttt{\splitchar{}} and \texttt{\joinchar{}}. Note that both the source and target sequence contain the same number of tokens.}\label{fig:pseudo-chars}
\vspace*{-.4cm}
\end{figure*}


\section{Related Work}\label{sec:related}

Historical text normalization is an ongoing research area that runs parallel to the entire development of the field of NLP. \textcite{bollmann_large-scale_2019} traces back text normalization back to the 1980s and gives a thorough overview up until 2019. Extending to current state-of-the-art methods, approaches to text normalization, both in historical and other contexts, can be subdivided into 1) context-agnostic type-based methods, and 2) context-aware sentence-based methods.

\subsection{Type-Based Methods}\label{sec:related-type}

Type-based methods approach text normalization by generating a normalized canonical replacement for each historical type variant occurring in the source text, and then replacing each occurrence of that particular historical type with its predicted normalized form. Historically, this approach was the most common one, in particular before the emergence of Transformer-based language models.

First approaches were mostly rule-based. These range from simple substitution lists (e.g., \cite{rayson_vard_2005} for English, \cite{gotscharek_towards_2011} for German) to edit rules on the character level in types, exploiting regularities in orthographic changes. For instance, the rule \graph{ey} $\rightarrow$ \graph{ei} would normalize the historic variant \graph{Freyheit} to its extant variant \graph{Freiheit}. 
These typically are even constrained on their surrounding context, e.g. only applying \graph{ss} $\rightarrow$ \graph{ß} when preceded by \graph{o} and succeeded by \graph{e} to normalize \graph{stossen} to \graph{stoßen} but not \graph{Risse} to *\graph{Riße}.
In principle, these edit rules can be either hand-crafted manually, following an intensional approach (such as in CAB; \cites{jurish_comparing_2010}{jurish_finite-state_2012}), or learned from a parallel corpus, following an extensional approach (such as in Norma; \cite{bollmann_semi-automatic_2012}).
These rules are not unambiguous: applying these edit rules to an unknown historic type does not yield a single normalization hypothesis, but possibly a multitude of normalization hypotheses.
As such, one common extension is the inclusion of a contemporary full-form lexicon and to only consider those hypotheses occurring in the lexicon. Furthermore, the edit operations can be weighted such that more plausible edit operations are preferred, with the intent that the correct hypothesis gets assigned more weight in total. This approach is been incorporated into CAB and Norma \parencites{jurish_comparing_2010}{bollmann_semi-automatic_2012}.

From a probabilistic perspective, the objective of type-based historic text normalization is to model the probability $p(\hat{w} \mid w)$ that the historic type $w$ has the normalized form $\hat{w}$. Again, note that the surrounding context of $w$ is not taken into account on type-level.
From this perspective, the task can be rendered as a translation task, and thus allows all typical methods of machine translations, learning to model the probability above from parallel corpora.
Methods range from statistical machine translation (SMT, e.g. type-based variant of Csmtiser; \cite{ljubesic_normalising_2016}), neural machine translation (NMT) via LSTMs \parencite{bollmann_learning_2017}, and—most recent—machine translation via sequence-to-sequence Transformers on character- and subword-level \parencites{tang_evaluation_2018}{wu_applying_2021}.
The first component of the model presented here directly follows this approach.

Due to the nature of type-based normalization, such systems cannot handle ambiguous cases or make use of the surrounding context. To alleviate this limitation, type-based text normalization models are usually augmented by incorporating a language model (\emph{language model} in the generic sense of an algorithm that assigns a probability to an utterance).  Since type-based models usually generate a distribution of plausible normalization hypotheses, 
one can evaluate all induced possible normalized candidate sentences using language models. \citeauthor{jurish_more_2010} (\citeyear{jurish_more_2010}; cf.~\citeyear{jurish_finite-state_2012}) uses hidden Markov models to incorporate bigram probabilities learned form a contemporary corpus. More general, \textcites{mitankin_approach_2014}{lusetti_encoder-decoder_2018}{sukhareva_context-aware_2020} rank the normalization hypotheses through an $n$-gram language model, combining both the likelihood returned by the type-based normalizer and the likelihood returned by language model.
The second component of the model presented here also follows this approach of including a language model to re-rank the moralization hypotheses.

\subsection{Sentence-Based Methods}\label{sec:related-sent}

One immediate limitation of the type-based perspective is that the surrounding context of a particular type occurring in the source text is not taken into account. As already pointed in the introduction, historic spellings can have more than one unique normalization, depending on their context in which they are used.
As such, contemporary approaches understand historic text normalization as a translation task on the sentence level, using language models to translate the entire sentence in its historic form to its extant form, using the full potential of current NLP methods.

The system \emph{ModFr} \parencite{bawden_automatic_2022} appears to be the first such publication that uses a sequence-to-sequence Transformer in an encoder-decoder fashion, trained from scratch using a parallel corpus to normalize historic spellings of French, translating on a sentence level.
A similar approach have been pursued in the system Transnormer by \textcite{bracke}, who seeks to design a normalization tool for the DTA, thus handles historic German c.~1500–1800, also makes use of the DTA EvalCorpus and follows a very similar objective as here.
Methodologically, unlike \citeauthor{bawden_automatic_2022}, \citeauthor{bracke} fine-tunes the sequence-to-sequence ByT5 Transformer model \texttt{google/byt5-small} (\SI{300}{\mega\nothing} parameters; \cite{xue2022byt5tokenfreefuturepretrained}), which has been pre-trained on a (non-parallel) multilingual corpus in a fill-mask objective. 
A similar setup has been pursued by \textcites{de_la_rosa_modernilproject_2022}{rubino_automatic_2024}, with the former normalizing Spanish golden age texts (c.~1600), and the latter fine-tuning the sequence-to-sequence Transformer model \texttt{facebook/m2m100\_418M} pre-trained on many-to-many multilingual translation, in order to normalize Middle French (c.~15th century).

Sentence-based approaches to text normalization were even pursued before the emergence of Transformers, albeit much less frequent. One example is the sentence-based variant of Csmtiser \parencite{ljubesic_normalising_2016}, that normalizes by taking entire sentences as input, where whitespace is treated as an individual pseudo-character of the alphabet, and then proceeding like the type-based variant.

\subsection{Discussion}

When type-based method were still the dominant approach, the general consensus on the different strategies (SMT, rule-based, NMT) appeared each strategy comes with their own strengths and weaknesses, and thus no model clearly outperforms the others \parencites{sang_clin27_2017}{hamalainen_normalizing_2018}{robertson_evaluating_2018}{bollmann_large-scale_2019}{bawden_automatic_2022}.
However, since the emergence of pre-trained LLMs, the type-based methods seem to be forgotten, with contemporary approaches to text normalization usually on a sentence-level training or fine-tuning a Transformer to contextually normalize inputs end-to-end \parencites{van_der_goot_multilexnorm_2021}{bawden_automatic_2022}{klamra_evaluating_2023}{rubino_automatic_2024}{hopton_modeling_2024}. One notable exception to this trend the direction taken by \textcites{partanen_dialect_2019}{kuparinen_dialect--standard_2023}, who took a middle ground and normalized not on word level, and not on sentence level, but on a level of a sliding window consisting of only 3 words, but again, no method clear stood out as superior.

Faced with the problem of data sparsity, some approaches also resorted to very large assistant language models such as GPT4 to generate synthetic parallel data, allowing them to fine-tune their normalization models \parencites{rubino_automatic_2024}{klamra_evaluating_2023}. In general, the capabilities of very large assistant language models (i.e., at least GPT3-sized) in text normalization are still underinverstigated.

Finally, recent efforts question the need for text normalization altogether. As outlined in the introduction, the primary motivation for text normalization had been that unnormalized texts are out-of-vocabulary for NLP systems, thus language understanding is severely inhibited. Empirical experiments further showed that BERT's performance is highly sensitive to noisy text like spelling mistakes, typos, dialects, etc.~\parencite{kumar_noisy_2020}.
Faced with this, the traditional approach is to set up pre-processing pipelines that try to minimize the noise of the input dataset before being processed by NLP models, similarly to the historical text normalization task presented here.

However, recent works start to question this workflow, and take a more fine-grained position towards noisy text, observing that some of the noisy text (which is usually removed) does convey critical signal and is required for certain downstream task \parencite{al_sharou_towards_2021}.
This also applies to dialects and nonstandard language varieties, and recent work investigate the actual necessity of orthographic normalization. For instance, in the case of Part-of-Speech Tagging in Occitan's dialects, it has been shown that a multilingual BERT model is capable of cross-dialect transfer learning, giving us a slight indication that “orthographic normalization may not be necessary when fine-tuning large, multilingual models” \parencite{hopton_modeling_2024}.
Whether this also applies to more complex CLS-related downstream tasks remains to be investigated.

\section{Dataset}\label{sec:dataset}

This section provides a detailed overview of the employed data sources.
We begin with a description of the source parallel corpus DTA EvalCorpus, which was already mentioned in the introduction, discussing its origins, structure, and alignment process.
Following this, we outline the process of constructing the specific dataset used in our experiments, detailing the selection criteria, pre-processing steps, and final composition.

\subsection{Source Corpus}
The DTA EvalCorpus is a XML-encoded token-aligned parallel corpus of multiple historical German text publications and their respective contemporary editions, with dates ranging from 1780–1901, constructed by \textcite{jurish_constructing_2013}. See Fig.~\ref{fig:dtaec-source} for an excerpt of one XML document. The corpus was intended to provide ground-truth data for an evaluation of the CAB normalizer. Historical texts were retrieved from the DTA, who digitized and OCR'd historical prints. The corresponding contemporary editions were taken from the (international) \emph{Project Gutenberg}\footnote{\url{https://gutenberg.org/}} and the \emph{TextGrid Digital Library}\footnote{\url{https://textgrid.de}}. For the Project Gutenberg, the origin of the digitized texts is unknown. The TextGrid Digital Library obtained their digital editions from \emph{Editura/Zeno.org}, who themselves digitized contemporary editions of the historic texts published by professional publishers. 

Thus for instance, \citeauthor{jurish_constructing_2013} token-aligned the digitization provided by the DTA of the novel's first print 
\begin{quote}\raggedright\linespread{1.0}\selectfont
Christian Dietrich Grabbe, \emph{Napoleon oder Die hundert Tage. Ein Drama in fünf Aufzügen}. Frankfurt am Main: Johann Christian Hermannsche Buchhandlung, 1831.\footnote{DTA Digital edition: \url{https://www.deutschestextarchiv.de/book/show/grabbe_napoleon_1831}}
\end{quote}
with the digitization of the contemporary edition in the TextGrid Digital Library
\begin{quote}\raggedright\linespread{1.0}\selectfont
    Christian Dietrich Grabbe, \emph{Napoleon oder Die hundert Tage. Ein Drama in fünf Aufzügen}, in \emph{Historisch-kritische Gesamtausgabe in sechs Bänden}, ed. Akademie der Wissenschaften Göttingen and Alfred Bergmann, vol. 2, 315–460. Emsdetten, Westphalen: Lechte, 1963.\footnote{TextGrid Digital edition: \url{https://hdl.handle.net/11858/00-1734-0000-0002-E63A-2}}
\end{quote}
which itself refers to the manuscript by Grabbe as their source reference, basis for above 1831 print, but reprinting in contemporary orthography. Bergmann does not explicitly state any normalization guidelines in this edition, except for a “modernization of orthography” without further details.
The alignment was first computed algorithmically, and then manually verified in a two-stage process by the annotators. First, on type-level, some frequent normalization pairs were analyzed and then (conservatively) accepted. The remaining token pairs not confirmed in the first stage were subsequently analyzed in the second stage, where the surrounding context was also taken into account.

The description of the DTA EvalCorpus also never makes explicit any precise notion of normalization (or “canonicalization”) to be followed in their correction process, except for rejecting changes in word order, lexical roots and morphosyntactic features.
Rather, the parallel corpus explicitly intends to draw on the linguistic intuitions of the contemporary edition's editors. Thus, the corpus captures the \emph{extensional} definition of text normalization (Sec.~\ref{sec:definition}): the correct normalization form is defined by the corpus resp.\@ the editors' decision.
In total, the corpus contains approximately \SI{5.4}{\mega\nothing} token pairs in 127 documents from 1780–1901, with 102 of them being literary documents, and the remaining documents primarily philosophical treatises.

\subsection{Dataset Construction}
Following the goal of historical text normalization in literary texts, a training and evaluation dataset is constructed. This starts from a subset of the 127 documents of the DTA EvalCorpus, selecting only literary documents tagged as novels, theater plays, novellas, letters, travel writing, fairy tales, and prose, which results in 85 documents. In particular, this removes the 12 works of poetry, but for these documents, we already have argued in Sec.~\ref{sec:definition} that the nature of text normalization may significantly differ from the one carried out on the other non-poetic documents.

\begin{figure*}[t]\centering
    \begin{subfigure}[t]{0.65\textwidth}
        \scriptsize\ttfamily\strut\par\vspace*{.4cm}
\includegraphics[scale=1.01]{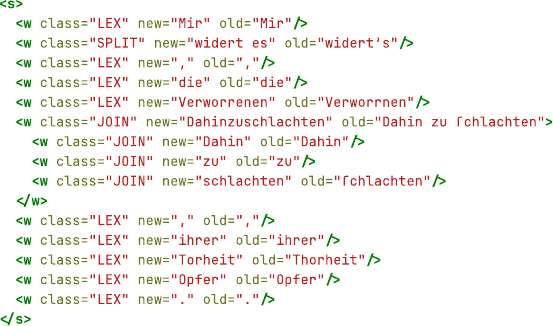}
\caption{Source code}\label{fig:dtaec-source}
    \end{subfigure}
    \begin{subfigure}[t]{.3\textwidth}
        \scriptsize\ttfamily\strut\par\vspace*{01pt}
        \begin{tabular}{ll}
            \toprule
            {\rmfamily\bfseries orig} & {\rmfamily\bfseries norm}\\
            \midrule
            Mir & Mir\\
            widert’s & widert\splitchar{}es\\
            , & ,\\
            die & die\\
            Verworrnen & Verworrenen\\
                & \\
            Dahin & Dahin\joinchar{}\\
            zu & zu\joinchar{}\\
            schlachten & schlachten\\
             & \\
            , & ,\\
            ihrer & ihrer\\
            Thorheit & Torheit\\
            Opfer & Opfer\\
            . & .\\
            \bottomrule
        \end{tabular}
    \vspace*{.6cm}
    \caption{\raggedright Corresponding section of the DTAEC dataset in the form of a token-aligned parallel corpus}
    \end{subfigure}
    \caption{Excerpt from the DTA EvalCorpus and the corresponding section in the transformed DTAEC dataset. Observe how the second pre-processing step already transliterated the \graph{\longs{}} in the “orig” column.}\label{fig:dtaeval}
\end{figure*}

The subset of documents was used to extract all token-aligned sentence pairs. Sentences that were manually annotated as “non-sentence-like unit” by the DTA EvalCorpus annotators were removed, as well as sentences that contained at least one token pair annotated as foreign material, printing-, \mbox{OCR-,} transcription error, or “unsuitable for inclusion in training material” (e.g., “tokenizer errors”, “extinct lexemes”, “hyphenated compounds”).

In the source corpus, instances where the token mapping is not 1:1 are encoded with special \texttt{JOIN} and \texttt{SPLIT} elements. As a first pre-processing step, the source corpus is brought into the 1:1 form described in Sec.~\ref{sec:definition} by performing a character-wise alignment on the source and target \emph{strings} (including whitespace) for every \texttt{JOIN} or \texttt{SPLIT} hunk,  using an Levenshtein alignment algorithm by \textcite{bawden_automatic_2022}. Pseudo-characters \texttt{\splitchar{}} and \texttt{\joinchar{}} are added to the target form accordingly. An example is shown in Fig.~\ref{fig:dtaeval}.
As a second pre-processing step, a trivial character transliteration is performed, replacing the long-s \graph{\longs{}} with the short \graph{s}, and replacing the Umlaut superscript-e with diaeresis, e.g. \graph{$\overset{\text{\tiny e}}{\text{a}}$} (U+0065 Latin Small Letter A, U+0364 Combining Latin Small Letter E) with \graph{ä} (U+00E4 Latin Small Letter A with Diaeresis). This completes the sampling process and pre-processing from the source corpus, resulting in \num{188687} sentences and \num{4007700} token pairs. Throughout this work, this dataset is referred to as the \emph{DTAEC dataset}.

To enable a typical machine-learning evaluation, the DTAEC is split into a training, a development, and a test split, approximating a 60–20–20 split. This split was stratified by time slices, and additionally was generated in such a way to keep all volumes of a multi-volume work (e.g., all three documents \emph{Der Nachsommer, erster Band, zweiter Band, dritter Band}) in the same split, in order to prevent test set contamination due to shared (work-specific) vocabulary. Quantitative descriptions of the splits are provided in Table~\ref{table:split}.
The full list of documents and their respective split is presented in Appendix~\ref{sec:split}.
The entire DTAEC dataset can be replicated from a copy of the (publicly available) DTA EvalCorpus.\footnote{\url{https://github.com/aehrm/hybrid_textnorm}}

\begin{table*}\centering
\sisetup{table-text-alignment=right}\small
\begin{tabular}{lS[table-format=11.0]S[table-format=7.0]rS[table-format=5.0]S[table-format=7.0]S[table-format=5.0]}
\toprule
{\bfseries Split} & {\bfseries \# Sentences} & \multicolumn{3}{c}{\bfseries \# Tokens} &  \multicolumn{2}{c}{\bfseries \# Types} \\ 
 \cmidrule(lr){3-5}\cmidrule(lr){6-7}
 & {} & {\bfseries Overall} & \bfseries Proportion & {\bfseries OOV} & {\bfseries Overall} & {\bfseries OOV} \\
\midrule
Train & 114618 & 2389022 & \SI{59.61}{\percent} & {—} & 101024 & {—} \\
Dev & 38425 & 917360 & \SI{22.89}{\percent}  & 29037 & 53222 & 19288 \\
Test & 35644 & 701318 &\SI{17.50}{\percent} & 26883 & 48084 & 16522 \\
\midrule
Total & 188687 & 4007700 & \SI{100.00}{\percent}  & {—} & 135092 & {—} \\
\bottomrule
\end{tabular}
\caption{Corpus statistics of the DTAEC splits. “OOV” refers to historical tokens resp.\@ types not present in the historical side of the training split.}\label{table:split}
\end{table*}

Just by descriptive statistics, we can already gain some insights into the nature of historic text normalization for this period and register. First, taking all splits together, most normalization pairs (\SI{96.251}{\percent}) are identity, meaning that performing the above transliteration often already yields the appropriate normalization. 
Furthermore, in general, text normalization for this dataset rarely involves spacing: on only \SI{0.461}{\percent} of token pairs, \texttt{\splitchar{}} and \texttt{\joinchar{}} are added in the target form.

More importantly, almost all text normalizations are unambiguous: from all \num{64703} historic types occurring at least twice, \SI{94}{\percent} had only a single corresponding extant type given by the DTAEC. Among the ambiguous cases, counted by type, most normalizations (\SI{65}{\percent}) are spacing-related or casing-related. In conclusion, ignoring the context and just taking the most frequent normalization form already yields the correct form in \SI{99.380}{\percent} of token pairs, and \SI{99.937}{\percent} when comparing spacing- and case-insensitive.

\section{Method}\label{sec:method}

This section describes the proposed hybrid system for historic text normalization on the established domain and dataset. The approach follows a two-stage process. In the first stage, an encoder-decoder Transformer model is used to normalize each historic type found in the text. This is done individually and without considering the context, resulting in a set of normalization hypotheses for each type.

As discussed, this ignores the surrounding context of the text. To address this, the second stage uses a language model trained to capture contemporary language features. For each sentence, this stage evaluates all possible sentence normalizations generated by the Cartesian product of all possible token normalizations from the previous model. This approach is loosely based on the work of \textcite{mitankin_approach_2014}, who also used a two-stage architecture to re-rank type normalization hypotheses.

One modification is made to this two-stage setup. Instead of calling the first stage on every historic type encountered, the system presented here follows the proposal of \textcite{robertson_evaluating_2018} and \textcite{bollmann_large-scale_2019}, who observe that type-based normalization models often struggle with memorizing the normalization pairs they were trained on. To address this, a simple substitution lexicon is set up. This lexicon lists the gold normalization form(s) for each word form encountered in the training set. The substitution lexicon can be easily obtained from the DTAEC training split by counting the frequency of historical-extant type pairs. For out-of-vocabulary (OOV) types, i.e., historical words not present in the lexicon and were not seen during training, the system falls back on the trained Type Transformer. This relies on the model's ability to generalize to unseen types.

Next Section~\ref{sec:type-transformer} describes the Type Transformer of the first stage, and Section~\ref{sec:language-model} explains the second stage, where a pre-trained LLM is used to re-rank the normalization hypotheses context-aware.


\subsection{Type-Based Type Transformer on Character Level}\label{sec:type-transformer}

\begin{table*}[t]\small\centering
    \begin{tabular}{lrrrrrr}
        \toprule
        \bfseries Model Size & \multicolumn{2}{c}{\bfseries \# Layers} & \# \bfseries Att. Heads& \multicolumn{2}{c}{\bfseries Dimensions} & \bfseries \# Parameters\\
        \cmidrule(lr){2-3} \cmidrule(lr){5-6}
        & Enc. & Dec. &  & Emb. & FFN\\
        \midrule
        S & 2 & 2 & 2 & 128 & 512 & $\approx$ \SI{1}{\mega\nothing}\\
        M & 4 & 4 & 4 & 256 & 1024 & $\approx$ \SI{7}{\mega\nothing}\\
        L & 6 & 6 & 8 & 512 & 2048 & $\approx$ \SI{45}{\mega\nothing} \\
        \bottomrule
    \end{tabular}
    \caption{Network sizes explored for the Transformer models. Model size L corresponds to the “base” Transformer size as per \textcite{vaswani_attention_2017}.}\label{tab:modelsize}
\end{table*}

\begin{table*}[t]\small\centering
\begin{tabular}{l@{\hskip 20pt}cc@{\hskip 20pt}cc@{\hskip 20pt}cc}
\toprule
\bfseries Vocab size & \multicolumn{2}{c}{123 (=\,alphabet)} & \multicolumn{2}{c}{200} & \multicolumn{2}{c}{500} \\
        \cmidrule(lr){2-3} \cmidrule(lr){4-5} \cmidrule(lr){6-7}
\bfseries WordAcc & Overall & OOV & Overall & OOV & Overall & OOV \\
\bfseries Model Size &  &  &  &  &  &  \\
\midrule
S & 92.4862 & 87.8318 & 92.7981 & 88.1895 & 92.6722 & 87.6918 \\
M & 95.4586 & \bfseries 90.9633 & 95.6165 & 90.7041 & 95.6071 & 90.4759 \\
L & 95.7255 & 90.7300 & \bfseries 95.9396 & 90.7870 & 95.7630 & 90.2478 \\
\bottomrule
\end{tabular}
\caption{Results of the hyper-parameter study, varying vocabulary size and model size, but keeping the learning rate (\num{1e-4}) and number of epochs (20) fixed. Reported numbers are the percentage of error-free output generations per type (not per token) on the development split, which was not part of the training. Column labeled “OOV” refers to the subset of types that are present in the development split, but not the training split. Highlighted numbers refer to the maximum overall accuracy resp.\@ OOV accuracy.}\label{tab:training-results}
\end{table*}

To perform the normalization on OOV types, i.e., not present in the substitution lexicon, an encoder-decoder model based on Transformers is employed, following a BART-style setup \parencite{lewis_bart_2019}. We will refer to this model as \emph{Type Transformer}. While, in general, other sequence-to-sequence models are conceivable (e.g., encoder-decoder LSTMs), Transformers come with the benefit of faster training, and prior research has shown that Transformers appear to either be comparable with LSTMs \parencite{bawden_automatic_2022}, or to even outperform them on sufficiently large  datasets \parencite{tang_evaluation_2018}.

For each historical type occurring in the training split of the DTAEC (i.e. also those that are covered by the lexicon), the model is trained to generate the corresponding normalized target form. If more than one normalization is present in the training split, then the most frequent normalization is chosen. As such, the number of training examples the model sees is identical with the number of types occurring in the historical side of the train split.
The inputs and outputs to the model are tokenized by the same Byte-Pair-Encoding (BPE) tokenizer \parencite{sennrich_neural_2016}, where the subword vocabulary (shared for both encoder and decoder) is derived from all tokens (historic and extant) of the training corpus.
Generation is performed with a beam search using four beams, which means that four output hypotheses are generated for each type resp.\@ input sequence.
Likelihood of predictions is calculated as usual as the joint probability of the output sequence through the chain rule, using the predicted conditional probabilities of a subword given the previous prefix, i.e., 
\begin{gather*} p_\text{TT}(\hat{c}_1\cdots \hat{c}_n\mid c_1\cdots c_m) \\= p_\text{TT}(\hat{c}_1\mid c_1\cdots c_m) \cdot p_\text{TT}(\hat{c}_2\mid \hat{c}_1, c_1\cdots c_m) \\ \qquad{}\cdot p_\text{TT}(\hat{c}_3 \mid \hat{c}_1\hat{c}_2, c_1\cdots c_m) \cdot \cdots\\
    = \prod_{i=1}^n p_\text{TT}(\hat{c}_i \mid \hat{c}_1\cdots \hat{c}_{i-1}, c_1\cdots c_m),
\end{gather*}
where $c_1\cdots c_m$ are the $m$ subwords of the historic type, and $\hat{c}_1, \dots$ the subwords of the predicted normalization, conditioned on $c_1\cdots c_m$.

A hyper-parameter search was conducted to maximize the accuracy (i.e., top predicted output sequence matches the gold output sequence) on the development set, again derived by taking the most frequent normalization for each historical type. In particular, different model sizes (see Table~\ref{tab:modelsize}) and different vocabulary sizes of the BPE tokenizer were evaluated. For all experiments, batch size is fixed to 8, learning rate to \num{1e-4} and the number of epochs to 20 epochs, since prior experiments has shown that these parameters yield the best results.
The results of this parameter search are summarized in Table~\ref{tab:training-results}. Increasing the vocabulary size above its theoretical minimum (the number of individual characters) allows the BPE tokenizer to form subword units spanning multiple characters. However, these do not appear to be beneficial to the model's performance. This confirms observations by \textcite{tang_evaluation_2018}. 
Considering the model size, we see that the overall performance increases with model size, but—as we would expect—generalizability to unseen types appears to degrade at sufficient model size, peaking at the M model size.
Thus, this work continues with the model size M for the Type Transformer, along with a character-based tokenizer.


\subsection{Re-Ranking through Sentence-Based Language Model}\label{sec:language-model}

Both the substitution lexicon (for in-vocab types) and the Type Transformer (for OOV types) come with the drawback that they cannot take in the context for a normalization, hence will make errors for ambiguous normalization cases. As outlined above, the second stage of the normalization system employs a LLM to choose more probable normalization hypotheses, by evaluating the possible sequences of modern candidate sentences induced by the normalizations generated by the previous stage. This allows finding the most probable sequence with respect to the modern language and the type-level normalization probabilities, also allowing the system to choose the correct normalization in ambiguous cases.

Given a historical form $w$, both the substitution lexicon and the Type Transformer do not only output a single normalization hypothesis, but a distribution of possible normalization hypotheses $\hat{w}$. 

For in-vocab types $w$ in the substitution lexicon, we can define the probability as the proportion of this normalization relative to all the observed ones, i.e.,
\[ p_\text{Lex}(\hat{w}\mid w) = \frac{f(w, \hat{w})}{\sum_{v} f(w, v)}, \]
where $f(w, v)$ is the frequency of historical-extant pairs $(w,v)$ in the DTAEC.
For an OOV type $w$, the Transformer generates four outputs $\hat{w}$ from $w$ by the beam search, with the probability $p_\text{TT}(\hat{w} \mid w)$ attributed by the Transformer's next-token prediction, as defined above.

Now, let $S=w_1w_2\cdots w_n$ be the historic sentence. By the lexicon resp.\@ the Type Transformer, we obtain, for each $w_i$, a set $H_i=\{ \hat{w}_i^{(1)}, \hat{w}_i^{(2)},\hat{w}_i^{(3)},\ldots  \}$ of the predicted normalization hypotheses (that is, having nonzero probability conditioned on $w_i$). We now can reformulate our normalization task to pick, for each $w_i$, a hypothesis $\hat{w}_i\in H_i$ such that the sentence $\hat{S}=\hat{w}_1\cdots\hat{w}_n$ is most plausible. To now define which sequence is more “plausible”, an approach similar to \textcite{mitankin_approach_2014} is followed by taking the joint probability of type normalizations and the overall likelihood of $\hat{S}$ with respect to modern language, with the latter estimated through a language model. 
Thus, formally, hyper-parameters $\alpha>0$ and $\beta>0$ are introduced and we are interesting in determining
\[ \hat{S} = \argmax_{\hspace*{.8cm}\mathclap{\hat{w}_1 \cdots \hat{w}_n \in H_1 \times\cdots\times H_n}\hspace*{0.8cm}} p_\text{LM}(\hat{w}_1 \cdots \hat{w}_n) \cdot \prod^n_{i=1} p_\text{type}(\hat{w}_i\mid w_i), \]
with
\[ p_\text{type}(\hat{w}_i \mid w_i) \propto \begin{cases} 
p_\text{TT}(\hat{w}_i\mid w_i)^{1/\alpha} & \text{if $w_i$ is OOV} \\ 
p_\text{Lex}(\hat{w}_i \mid w_i)^{1/\beta} & \text{otherwise}. \end{cases} \]
Here, $p_\text{LM}$ denotes a probability distribution estimated by a language model, modeling the likelihood of (full) utterance $\hat{w}_1\cdots\hat{w}_n$. The hyper-parameters $\alpha,\beta>0$ control how much relevance is given to the orthographic normalization probabilities  as opposed to the likelihood of the full sentence as estimated by the language model.

In that sense, $\alpha$ and $\beta$ induce a spectrum between the impact of the spelling variations and that of the general language features: With $\alpha, \beta\to 0$, the system degenerates to taking the most plausible type translation, isolated for each historic token. With $\alpha, \beta\to \infty$, the system ignores all translation likelihoods and only focuses on maximizing the language-model-likelihood of the full sentence from the given normalization hypotheses $H_1, H_2, \ldots$, without taking into account their probabilities conditioned on their historic form.
Furthermore, differences between $\alpha$ and $\beta$ amount to assigning different certainties to the estimated probability distributions. For instance, setting $1=\alpha<\beta$ would leave the distribution of the Type Transformer unchanged, but would “smooth off” the empirical frequency distribution of the lexicon.


While in principle it is possible to evaluate each candidate sentence and calculate its plausibility as defined above, in order to find the best one $\hat{S}$, this approach is generally computationally infeasible since the number of possible candidate sentences grows exponentially with sentence length.
Instead, the system resorts to a beam search heuristic through the tree induced by the token hypotheses $H_i$, calculating plausibility for a beam not using the the full sentence, but only for the prefix of the sentence up until the end of the beam. 

As language model, this systems uses a German off-the-shelf GPT2 \texttt{dbmdz/german-gpt2}\footnote{\url{https://github.com/stefan-it/german-gpt2}} \parencite{stefan_schweter_2020_4275046}, trained on a Wikipedia dump, EU Bookshop corpus, Open Subtitles, CommonCrawl, ParaCrawl and News Crawl (\SI{16}{\giga\byte} data; \SI{137}{\mega\nothing} parameters). To derive the likelihood of a full sequence from GPT's causal language modeling objective, the likelihoods of the next-token predictions are accumulated using the chain rule, like in above with the Type Transformer, but in particular on the GPT-tokenized sequence $v_1\cdots v_n$, and after detokenizing and removing the pseudo-characters \texttt{\splitchar{}} and \texttt{\joinchar{}} accordingly.
Beam search was performed with four beams. Prior experimentation on the development split of the dataset yield best results with hyper-parameters $\alpha=\beta=0.5$, though it should be highlighted that these hyper-parameters are dependent on the text domain and on the language model employed.

While better German BERT models exist (in intrinsic and extrinsic performance, and with the training data closer to the historical literary domain), due to its fill-mask training objective, it cannot be easily used to estimate likelihoods of full sequences like with causal LLMs such as GPT. Nevertheless, more choices of potential language models should be investigated in future work.


\section{Evaluation}\label{sec:evaluation}

Having described the system, the following section performs an evaluation of the proposed hybrid system against several other baseline systems, including CAB, and several ablated variants of the hybrid system. The respective systems and experimental setup will be described in Sec.~\ref{sec:systems}, followed by a quantitative evaluation and comparison, as measured by accuracy on the DTAEC dataset in Sec.~\ref{sec:comparison}. Next to the full accuracy, a more fine-grained analysis will be pursued, on the one hand through (algorithmically) categorizing the systems' errors in Sec.~\ref{sec:error-categories}, and a qualitative error analysis of the proposed hybrid system in Sec.~\ref{sec:error-qualitative}. Finally, in Sec.~\ref{sec:memorization}, to validate the measured performance of the hybrid system, a test on memorization will be conducted to rule out the possibility of test set contamination due to the hybrid system's GPT2 model trained on crawled internet data.

\subsection{Systems}\label{sec:systems}

Next to the full proposed hybrid system, we want to also assess the importance of the individual components of the hybrid text normalization system, three different variants are compared: \textbf{Type Transformer} refers to only the type Transformer of Sec.~\ref{sec:type-transformer} skipping the substitution lexicon and the language model re-ranking, and thus processing every type in isolation.
The second variant \textbf{Type Transformer~+Lex} adds the substitution lexicon but skips the language model. Thus, in comparison to the full system, this variant degenerates to only taking the most plausible type translation, isolated for each historic token, which is equivalent to setting $\alpha, \beta\to 0$.
The third variant \textbf{Type Transformer~+Lex+LM} refers to the full proposed hybrid variant, with the empirically determined hyper-parameter values $\alpha=\beta=0.5$.

\paragraph{Baselines.}

As simple baseline systems, the \textbf{Identity} and \textbf{Lexicon} baselines will be considered in the evaluation. The first system only copies the historical word forms without making any modifications. Note that inputs from the DTAEC are already transliterated, e.g., replacing \graph{\longs{}} with \graph{s}. The second system replaces each historical type with its most frequent normalization in the training set. This amounts to only using the substitution lexicon, or hyper-parameters  $\alpha\to\infty, \beta\to 0$.

As a diagnostic measure, the \textbf{Best Theoretical Type Normalization} is added. This system replaces each type with its most frequent normalization in the \emph{test set}, where the evaluation is run. Since the frequencies are derived from the gold test set, this is only a theoretical system. However, the performance achieved by this system is the maximum attainable performance for a purely type-based normalization system, which by definition ignores the surrounding context.

\paragraph{CAB/DTA.}
The system described here is compared against the CAB as used in the DTA. Unlike the other models which have been trained and executed locally, the CAB/DTA is not publicly available, and therefore is only invoked through the API access to the “production” instance exposed by the DTA.\footnote{\url{https://deutschestextarchiv.de/public/cab/}; see also \url{https://www.deutschestextarchiv.de/doku/software\#cab}} 
Starting from Jurish's system, the CAB/DTA has added an optional “exception lexicon”, making corrections to the system using a database of exceptions, extracted from the DTA EvalCorpus, but of unknown size and extent \parencite[cf.][]{jurish_using_2015}. As such, the “default” CAB/DTA normalizer is likely memorizing parts of the DTA EvalCorpus, possibly even from the test split considered here. Due to this data leakage, the performance of the CAB/DTA normalizer cannot be compared with the other systems, and therefore it is only reported for informational reasons.

As an alternative, a second evaluation is separately performed which disables the exception lexicon through the API options (\textbf{CAB/DTA w/o Exlex}). Unlike the above system with the exception lexicon, the original system was not trained at all, except for the $n$-gram language model on out-of-domain contemporary textual data. In this case, the measured performance can be compared to the other (trained) systems.

\paragraph{Norma.}

The normalization system \emph{Norma}\footnote{\url{https://github.com/comphist/norma/}} by \textcite{bollmann_semi-automatic_2012} is also compared against. It works type-based by combining a substitution lists with weighted edit rules on the character-level, learned from a parallel corpus. During inference, for OOV words, the prediction is chosen from a full-form lexicon with smallest edit weight from the historical type. In the experiments, the edit rules are learned form the training set of the DTAEC. Additionally, to ensure comparable performance, a general-purpose contemporary German word list (unrelated to the DTAEC) is supplied as the full-form lexicon.  

\paragraph{Csmtiser.}

As representant of a normalization system based on SMT, the evaluation also covers \emph{Csmtiser}\footnote{\url{https://github.com/clarinsi/csmtiser}} by \citeauthor{ljubesic_normalising_2016} (\citeyear{ljubesic_normalising_2016}; cf.~\cite{scherrer_automatic_2016}), employing the general-purpose SMT decoder \emph{Moses} on character-level. This system comes in two variants: the first one is type-based, and decodes historic types into its most probable extant form on character-level (\textbf{Csmtiser\textsubscript{Type}}), reweighted by an $n$-character language model trained on contemporary word forms. Like in the experiments by \textcite{bollmann_large-scale_2019}, the decoder is trained on the parallel training split of the DTAEC, and the language model on the target side of the training split.

The second variant works sentence-based and context-aware, by translating historic sentences, again on character-level (\textbf{Csmtiser\textsubscript{Sent}}), treating whitespace as a pseudo-character in the alphabet. Similarly, the decoding hypotheses are reweighted by an $n$-character language model, but which was trained on full contemporary sentences.  Again, for the experiments both the decoder and the language model are trained on the training split of the DTAEC, but in this variant sentence per sentence, with the sentences generated by detokenizing the DTAEC.

To align the output sentences (i.e., strings) of the Csmtiser\textsubscript{Sent} model back to the original tokenization of the DTAEC, the alignment technique by \textcite{bawden_automatic_2022} was utilized. This involved searching for the optimal sequence of (weighted) Levenshtein edit operations.

\paragraph{Transnormer.}

Finally, the  sequence-to-sequence encoder-decoder Transformer model Transnormer\footnote{\url{https://github.com/ybracke/transnormer}} by \textcite{bracke} is evaluated, which was already sketched in Sec.~\ref{sec:related-sent}: given a sentence in historic form, the model generates the respective sentence in contemporary orthography.  Transnormer is trained not from scratch but fine-tuned from the foundational model \texttt{google/byt5-small}. For the experiments, the same foundational model was fine-tuned, sentence by sentence, on the training split of the DTAEC, again detokenized, with the same hyper-parameters as Bracke (four gradient accumulation steps, each of batch size 2; 9 epochs; learning rate \num{5e-5}).
Alignment to the tokenization of the DTAEC was again done by Levenshtein edit search, like above.

\subsection{Evaluation and Comparison}\label{sec:comparison}

The different models will primarily be evaluated by \emph{word accuracy} (WordAcc) over all tokens in the DTAEC, which checks for each historical-extant pair $(w, \tilde{w})$ if the predicted normalization $\hat{w}$ is identical with the gold normalized form $\tilde{w}$.
Throughout this section, punctuation tokens will be ignored.
As recommended by \textcite{robertson_evaluating_2018}, not only word accuracy on all tokens is reported (Overall), but also word accuracy on the subset of pairs $(w,\tilde{w})$ where $w$ was present in the training set vocabulary (Invocab) resp.\@ where it was not (OOV). This is motivated by the fact that OOV Word Accuracy is the only metric that accurately reflects the systems' capacity to generalize to previously unseen tokens. In contrast, the overall Word Accuracy is confounded by the systems' ability to memorize normalizations that have been encountered before.

Table~\ref{tab:results} shows the results of this evaluation. Note that approximately \SI{3.833}{\percent} of tokens in the test set are OOV (i.e., not seen in the training set).
The table additionally marks the \emph{relative error} which flips and rescales the WordAcc columns, individually for the number of overall errors and the OOV errors. Formally, we define
\[ \text{RelativeError}(x) = \frac{1-\text{WordAcc}(x)}{1-\text{WordAcc(Identity)}} \]
for system $x$.
In other words, the error count of the Identity baseline is \emph{defined} as \SI{100}{\percent} relative error units, and the error count of the other systems is scaled accordingly. Thus, the relative error percentage spans the range of doing no normalization at all (\SI{100}{\percent}) and perfect accuracy (\SI{0}{\percent}).

Concerning the overall word accuracy, we see that the Transformer-based models perform best on this dataset, all making fewer errors than CAB/DTA (5) and the other baseline systems (6–8). Among the Transformer-based models, the proposed hybrid model (12) and the sentence-based Transnormer (9) having largely equal performance, both making half as much errors as the CAB/DTA system, with the hybrid system slightly better on in-vocab tokens, and Transnormer slightly better on OOV tokens. Concerning the research question raised in the introduction, the hybrid model appears to be unable to make better use of the training material than sentence-based Transformers. Both approaches lead to the same performance.

\begin{table*}[t]\small\centering
    \begin{tabular}{l@{\hspace*{5pt}}lrrrrr}
\toprule
&\bfseries System  & \multicolumn{3}{c}{\bfseries WordAcc $\uparrow$} & \multicolumn{2}{c}{\bfseries Rel. Error $\downarrow$} \\
\cmidrule(lr){3-5} \cmidrule(lr){6-7}
&  & Overall & Invocab & OOV & Overall & OOV  \\
\midrule
\rlap{\emph{Baselines}} & & &  \\
(1) & Identity & 96.514 & 97.017 & 83.912 & \underline{100.000} & \underline{100.000} \\
(2) & Lexicon & 98.881 & 99.477 & 83.912 & 32.109 & 100.000 \\
(3) & CAB/DTA* & 98.948 & 99.113 & 94.818 & 30.178 & 32.208 \\
(4) & Best Theoret.\@ Type* & 99.548 & 99.534 & 99.896 & 12.968 & 0.647 \\
\midrule
\rlap{\emph{Rule-based/Statistical}} & & & & \\
(5) & CAB/DTA w/o Exlex & 98.072 & 98.502 & 87.278 & 55.320 & 79.075 \\
(6) & Norma & 96.834 & 99.477 & 30.521 & 90.824 & 431.861 \\
(7) & Csmtiser\textsubscript{Type} & 98.940 & 99.321 & 89.369 & 30.415 & 66.081 \\
(8) & Csmtiser\textsubscript{Sent} & 98.928 & 99.317 & 89.160 & 30.763 & 67.376 \\
\midrule
\rlap{\emph{Transformer-based}} & & & & \\
(9) & Transnormer & 99.146 & 99.429 & \bfseries 92.032 & 24.500 & \bfseries 49.528 \\
(10)  & Type Transformer & 98.445 & 98.788 & 89.823 & 44.627 & 63.260 \\
(11) & Type Transformer +Lex & 99.107 & 99.477 & 89.823 & 25.609 & 63.260 \\
(12) & Type Transformer +Lex+LM & \bfseries 99.194 & \bfseries 99.493 & 91.701 & \bfseries 23.117 & 51.584 \\
\bottomrule
\end{tabular}
\caption{Accuracies on the DTAEC test split. Reported numbers are all percentages. Column “Invocab” refers to the subset of historical-extant pairs where the historical form is present in the training split, columns “OOV” to the subset of pairs where the historical form is not present, i.e., forms never seen during training. Relative error are two abstract unit of number of errors, where \SI{100}{\percent} is defined as the number of errors the Identity baseline makes overall resp.\@ on the OOV subset. Systems marked with $\ast$ do not follow the train/test split defined by the dataset, and thus their results do not reflect in-production performances. Underlined number are not the result of measurement but of definition. Highlighted number are best per column, excluding systems marked with $\ast$. Remember that CAB/DTA without Exlex is purely rule-based and is not trained with any data; here the division between “Invocab” and “OOV” is maintained purely for diagnostic reasons.}\label{tab:results}
\end{table*}

One central research question of this report was whether a type-based normalizer can benefit from a language model re-ranking. Comparing the variant without the language model (11) and the full hybrid model with the language model (12) confirms this, in particular for the OOV performance. The results suggest that on the one hand it is indispensable to add contextualized information for accurate normalization, but on the other hand the specific method (sequence-to-sequence vs.\@ re-ranking) plays only a minor role.

Furthermore, concerning the type-based approach pursued here, we see that the substitution lexicon is a crucial component to achieve state-of-the-art accuracies. 
This can be verified by comparing the accuracy of the Type Transformer alone (10) vs.\@ the variant falling back on the lexicon but without any re-ranking (11), where the overall word accuracy is significantly lifted.
This also confirms the empirical trend that that for \emph{seen} tokens, the lexicon lookup baseline (2) is extremely powerful and hard to beat \parencites{robertson_evaluating_2018}{bollmann_large-scale_2019}. Additionally, this also adds evidence to the hypothesis that learning normalization patterns is actually very hard, at least for Transformers: even on instances seen during the training, the Type Transformer (10) cannot reproduce them all accurately, and even worse than the type-level SMT-based Csmtiser (7).

In comparison to the best theoretic type-based accuracy (4), we see that all systems are struggling with generalizing the normalization task on OOV words. In theory, an OOV word accuracy of \SI{99.8}{\percent} is attainable while ignoring the context, only through normalizing type per type. But even the proposed hybrid system (12) and the sentence-based Transnormer (9) come only to around \SI{92}{\percent}. Comparing with the identity baseline (1, i.e., leaving as-is), both the hybrid system and the sentence-based Transnormer only reduces error count by half, showing that there is still a long way ahead for generalizing text normalization. Analogously, these findings may also indicate that text normalization, while not necessarily context-sensitive, is actually highly irregular and unpredictable.

Surprisingly, the relatively lightweight sentence-based SMT system Csmtiser\textsubscript{Sent} (8) performs relatively good, both on in-vocab tokens and OOV tokens, approaching the performance of Transformer-based systems, but requiring only moderate CPU resources. This is consistent with previous research \parencites{bollmann_large-scale_2019}{sukhareva_context-aware_2020}{kuparinen_dialect--standard_2023}, and again hints at the complexity of the task in the $>\!\SI{98}{\percent}$ range, where even sophisticated systems make little improvements.
Taken together, there appears to be a performance ceiling, at least in the DTAEC dataset, that is particularly hard to beat. 

\begin{table*}\small\centering
\setlength{\tabcolsep}{3pt}
\begin{tabular}{l@{\hspace*{5pt}}lrrrrrrr}
    \toprule
    & \bfseries System& \bfseries Overall& \multicolumn{3}{c}{\bfseries Invocab} & \multicolumn{3}{c}{\bfseries OOV} \\
    \cmidrule(lr){4-6} \cmidrule(lr){7-9}
    &  &  & Spacing & Casing & Character & Spacing & Casing & Character \\
    \midrule
\rlap{\emph{Baselines/Rule-based}} & & & & & & & & \\
(1) & Identity & \underline{100.000} & 13.635 & 4.533 & 68.783 & 1.051 & 0.601 & 17.051 \\
(2) & Lexicon & 32.109 & 9.519 & \bfseries 3.498 & \bfseries 1.837 & 1.051 & 0.601 & 17.051 \\
(4)  & Best Theoret.\@ Type* & 12.968 & 9.213 & 3.076 & 0.884 & 0.008 & 0.029 & 0.078 \\
    \midrule
\rlap{\emph{Statistical/Transformer-based}} & & & & & & & & \\
(7)  & Csmtiser\textsubscript{Type} & 30.415 & 9.949 & 4.214 & 5.298 & 0.618 & 0.667 & 10.910 \\
(9)  & Transnormer & 24.500 & \bfseries 7.822 & 4.017 & 4.537 & \bfseries 0.425 & \bfseries 0.560 & \bfseries 8.124 \\
(10)  & Type Transformer& 44.627 & 9.851 & 4.439 & 19.693 & 0.466 & 0.610  & 10.501\\
(11) & Type Transformer +Lex& 25.609 & 9.519 & \bfseries 3.498 & \bfseries 1.837 & 0.466 & 0.610 & 10.501 \\
(12)& Type Transformer +Lex+LM & \bfseries 23.117 & 8.656 & 3.923 & \bfseries 1.837 & 0.454 & 0.581 & 8.394 \\
    \bottomrule
\end{tabular}
\caption{Categorization of the errors made by select systems. Reported numbers are all percentages of relative error, where \SI{100}{\percent} is defined as the number of errors the Identity baseline makes in total. In this table, the “Invocab” and “OOV” columns are all of the same unit, i.e., relative to the \emph{total} number of errors of the Identity baseline, both in-vocab and OOV. Note that the categorization is not disjoint, hence the sum per row may add to more than the value marked as “Overall”. Highlighted numbers are lowest per column, excluding the hypothetical best-type baseline.}\label{tab:errors}
\end{table*}

\subsection{Error Categorization}\label{sec:error-categories}
Table~\ref{tab:errors} gives a breakdown on the error types selected systems makes during inferences. We can characterize the errors by analyzing which Levenshtein edit operations are necessary to get from the predicted token $\hat{w}$ to the true extant form $\tilde{w}$. If some operation involves a spacing pseudo-character \texttt{\splitchar{}} or \texttt{\joinchar{}}, it is marked as \emph{spacing error}; if some operation involves swapping a lower-case with the respective upper-case character, it is marked as \emph{casing error}; other operations (i.e., those involving new characters modulo casing) are marked as \emph{character error}.

Having identified the slim improvement of the full hybrid system (12) in comparison to the  one skipping the language model re-ranking (11) in the previous section, this categorization allows us to closer investigate the effect size of the re-ranking. The improvement comes through two capabilities of the language model: First, it reduces spacing errors on ambiguous in-vocabulary tokens, taking into consideration the surrounding context, precisely as we expected by our research hypothesis.

Second, we wee an improvement in the character errors in the OOV tokens, i.e., where the top prediction by the Type Transformer itself does differ by more than spacing and character case. Here, the language model rather finds the most fitting normalization from the Transformer's hypotheses, taking advantage the context.  However, unlike our research hypothesis, this is not due any ambiguity of the OOV tokens: almost all normalizations are unambiguous in the OOV test set, and instead we see how the language model (just) makes use of its larger linguistic knowledge to distinguish between the “plausible” and “unplausible” normalization hypotheses by the Type Transformer.
\begin{figure*}[t]\small
\begin{enumerate}[leftmargin=08em,label=(\arabic*)]
    \item \begin{tabular}[t]{ll}
        Source & \emph{Sie keuchten, und er \underline{schnob} und zitterte [...]}\\
        Target & \emph{Sie keuchten, und er \underline{schniefte} und zitterte [...]}\\
        Pred. & \emph{Sie keuchten, und er *\underline{schnob} und zitterte [...]}
    \end{tabular}
\item \begin{tabular}[t]{ll}
        Source & \emph{[...] in \underline{weißseidnen} Schuhen [...]}\\
        Target & \emph{[...] in \underline{weißseidenen} Schuhen [...]}\\
        Pred. & \emph{[...] in *\underline{weißenden} Schuhen [...]}
    \end{tabular}
\item \begin{tabular}[t]{ll}
        Source & \emph{[...] Menge von Ringen, die \underline{glitzten} und glimmerten [...]}\\
        Target & \emph{[...] Menge von Ringen, die \underline{glitzerten} und glimmerten [...]}\\
        Pred. & \emph{[...] Menge von Ringen, die *\underline{glitzten} und glimmerten [...]}
    \end{tabular}
\item \begin{tabular}[t]{ll}
        Source & \emph{[...] die \underline{Baare} der Kirche [...]}\\
        Target & \emph{[...] die \underline{Bahre} der Kirche [...]}\\
        Pred. & \emph{[...] die *\underline{Bare} der Kirche [...]}
    \end{tabular}
\end{enumerate}\caption{Selection of character errors made by the hybrid system on the DTAEC test split.}\label{fig:errors}
\end{figure*}

In comparison to the sentence-based Transnormer (9), in previous section we have already seen  Transnormer slightly outperforming the hybrid system (12) on OOV tokens. The detailed categorization here now shows that this is the case across all categories. For in-vocabulary tokens however, the hybrid systems appears to better handle character variations, though this is the result of the substitution lexicon, not of the language model re-ranking. On the other hand, Transnormer makes fewer spacing-related errors, which could be the result of its tokenizer-free end-to-end approach to text normalization. 

\subsection{Qualitative Error Analysis}\label{sec:error-qualitative}
In addition to the quantitative analysis presented, a small qualitative evaluation was also conducted. This involved a random subset of 200 erroneously normalized tokens on the test set (\SI{3.5}{\percent} of all token errors) made by the hybrid system (12). Overall, it was observed that in certain cases, the system produced normalizations that could be intuitively interpreted as appropriate normalizations. This will be demonstrated for each of the three error categories mentioned above: spacing errors, casing errors, and character errors.

Considering the spacing errors, in the majority of cases we can observe phenomena where, speaking from an intuitive point of view, the system's prediction is equally valid as the gold normalization provided by the DTAEC. For instance, in the case \emph{Alle Tage ging sie  \underline{dreimal} / \underline{drei Mal} in die Kirche}, or \emph{[...] in der wenigen Hoffnung, ihn bald \underline{wiederzusehen} / \underline{wieder zu sehen}}, where the variant after the dash is the gold variant.
A very strong disagreement between the predictions and the gold standard is on cases of \emph{so/wie/zu} + adjective, where the predictions prefer the separated forms, whereas the gold standard prefers the compounded form: \emph{es war \underline{zuviel} / \underline{zu viel} für sie}; \emph{[...] dass er bald doppelt \underline{soviel} / \underline{so viel} Schulden hatte}. In fact, this difference likely arises due to prescriptive language changes due to the 1997 German spelling reform. These prescribe the separated form (except when \emph{soviel} is used as conjunction; this is not the case in the sampled errors), but the reform took effect only after the publication of the editions. Instead, we may see the effect of the system's GPT2 language model (cf. Sec.~\ref{sec:language-model}) preferring the reformed orthography, which could be explained by the fact that the \texttt{dbmdz/german-gpt2} model is mostly trained on post-1997 crawled internet texts. This may also be the reason why the system consistently prefers separating word forms, like in \emph{bis der Gevatter Fallmeister \underline{vorüber kam}}; \emph{was ihm \underline{wohl tat}}; \emph{dicht \underline{in einander gepfropft}} (all of which prescribed as “incorrect” by current spelling regulations).

On the level of casing errors, again we can see that some normalizations marked as error can considered acceptable, due to some flexibility in casing after colons etc. In particular, the DTAEC itself is inconsistent on whether to start direct quotation capitalized, for instance. On the remaining cases, we can observe genuine errors arising from incorrect handling of adjectives used as nouns (\emph{alles \underline{Bisherige}}), formal pronouns (\emph{\underline{Ihre} Hochzeit}), and confounding with spacing errors (\emph{wenn man gut \underline{haushält} / \underline{Haus hält} damit}).

The final category of character error reveals the capabilities and limitations of the system's ability to generalize, as well as the limitations and ambiguities of the DTAEC. The hybrid system appears to struggle, particularly when dealing with OOV tokens, as depicted in Fig.~\ref{fig:errors}.

\begin{table*}[t]\small\centering
    \begin{tabular}{lr}
        \toprule
        \bfseries System configuration & \bfseries WordAcc OOV \\
        \midrule
        Type Transformer Recall@4 & 96.574\\
        \midrule
        Type Transformer ($\alpha,\beta\to 0$) & 89.823\\
        Type Transformer +LM, inferred hypothesis prob. ($\alpha=\beta=0.5$) & 91.701\\
        Type Transformer +LM, flat hypothesis prob. ($\alpha, \beta\to\infty$) & 80.054\\
        \bottomrule
    \end{tabular}
\caption{OOV word accuracy on the test set, for select variants of the proposed hybrid model. All numbers are percentages. The Transformer Recall@4 baseline is the proportion of normalizations where the gold normalization is part of the four hypotheses generated by the Type Transformer.}\label{tab:ablation}
\end{table*}

This is particularly pronounced in named entities, unmodified \emph{Wüllersdorf} becomes \emph{Willersdorf} in the prediction, \emph{Katthagen} to \emph{Kattagen}, \emph{Meister Frey} to \emph{Meister Frei}, the genitive-\emph{en} of \emph{Selindens} is not normalized to (gold) \emph{Selindes}, etc. However, we can also observe that the normalization standard for named entities varies across the dataset: original and predicted forms \emph{Niepeguk}, \emph{Mietz} get gold-normalized to \emph{Niepeguck}, \emph{Mieze}, but, e.g., original and gold-normalized \emph{Pankrazius} gets predicted as \emph{Pankratius}, which could be equally valid. A similar pattern applies to foreign words, where the predictions can be reasonably treated as correct: \emph{Peruque, Combinationen} is predicted as \emph{Perücke, Kombinationen}, or, symmetric, \emph{Obristenlieutenant} is maintained in the prediction instead of gold \emph{Obristenleutnant}. 
Additionally, there are sometimes particularly radical normalizations that were not picked up by the system: \emph{geahndet} to \emph{geahnt}, \emph{Schluft} to \emph{Schlucht}, \emph{Ältern} to \emph{Älteren} (instead of predicted and appropriate \emph{Eltern}).

In total, we can conclude that for one, there is still considerate room for improvements of the system's normalizing capabilities. But for another, the systems discussed here are already so good that a proper quantitative evaluation by (just) comparing predictions with the target side of the DTAEC parallel corpus comes to its limits: normalization systems frequently generate “incorrect” forms that are, in an intuitive sense, entirely appropriate as normalization. Furthermore, the dataset itself is inconsistent in some cases, possibly due to inter-editor differences.

\subsection{Test on Memorization}\label{sec:memorization}

The GPT2 language model employed in the full hybrid normalization model is pre-trained on a large crawled dataset, for which the precise content cannot be reliably reconstructed. Thus, even in the experiment presented here, we are faced with the  problem of test set contamination, as is usually the case when working with pre-trained LLMs. In particular, it is likely that the full contents of the works compromising the DTAEC test set are present in the GPT2 training set. This opens the possibility that the improvement of the language model re-ranking above is not a consequence of the language model's general linguistic ability, but rather its ability to reproduce works memorized.
As such, the results presented above may be overestimating the actual performance on truly unseen data.

While only training a GPT2 language model from scratch using a carefully constructed training set can completely rule out any test set contamination, we can at least, for the use case here, probe the language model's ability to reproduce the test set texts, in order to better interpret the results presented above.
Intuitively, we could just prompt the language model to complete sentences from the test set and assess its accuracy, but in order to provide a comparative analysis, I opt to compare the performance of the full hybrid normalization system with an ablated system that ignores the probabilities assigned to the normalization hypotheses. Thus, for instance, the full system needs to pick the correct normalization “\emph{Wir finden wohl keine [vergnügteren/vergnügtern/vergnügen/\linebreak[1]Vergnügen] Tage}” while taking into account the prior probabilities of \SI{80}{\percent}, \SI{10}{\percent}, \SI{9}{\percent}, \SI{1}{\percent} assigned to the four hypotheses derived by the Transformer from the historical form \emph{vergnügtern}, whereas for the ablated system, prior probabilities are flat (\SI{25}{\percent} each). This amounts to assigning parameters $\alpha, \beta\to\infty$ in the beam search.
Table~\ref{tab:ablation} shows the respective OOV word accuracy when evaluated on the test set. Additionally, overall recall of the hypothesis generation, i.e., the maximum attainable accuracy, is reported.
Here, the ablated system performed worst.
Under the assumption that the GPT2 model can indeed fully reproduce memorized text from the surrounding context, we should have observed the ablated system with the flat probabilities to accurately identify the correct normalization from the four given hypotheses, which would have achieved accuracies close to the \SI{96}{\percent} hypothetical maximum.
But instead, the ablated system performs significantly worse than even the Type Transformer predictions, which rely on the historical type forms only, and do not have any information about the surrounding context.
\looseness=1

In total, the experiment cannot completely rule out the possibility that, due to test set contamination, performance of the hybrid model on truly unseen data could be worse than measured in previous section.  However, the experiment admits the interpretation that even if there would be a test set contamination, the ability of the GPT2 language model to reproduce memorized information is limited. Instead, it is the combination of both textual context and of the historical source word form that gives the hybrid system its performance.

\section{Conclusion and Open Questions}\label{sec:conclusion}

Having identified the need for Digital Humanities to normalize orthography in historic documents to its contemporary form, this work aimed to develop a system that can automatically perform this orthographic normalization. In particular, the focus was on German literary language of the 19th century, as this normalization is necessary for Computational Literary Studies and Computational Stylistics to process their texts of interest in downstream NLP tasks, for which a dataset was established from a pre-existing parallel corpus “DTA EvalCorpus” by \textcite{jurish_constructing_2013}. This corpus aligned historical texts by the DTA with their respective contemporary editions, which come with a (manual) normalized orthography.

Some work concerning this type of normalization exists, in particular using a sentence-based Transformer system based on pre-trained LLMs, modeling normalization as translation task. In contrast, this report instead presented a smaller type of hybrid system, combining type-based and sentence-based approaches, following the research question of whether type-based models can make more effective use of the parallel corpora used to train such systems, thus outperforming other (e.g., sentence-based) approaches.
The system is hybrid in the sense that it consists of 1) a character-based encoder-decoder model trained from scratch to normalize, in isolation, each type in historic form to its extant form, and 2) an off-the-shelf pre-trained causal LLM that takes the results of the first component to normalize sentence by sentence, making use of the surrounding context to re-rank and select the correct normalization, particularly in ambiguous cases.

A quantitative evaluation against baseline systems indicated that, overall, the proposed hybrid model performed best in this particular parallel corpus, significantly better than the de-facto standard CAB normalizer, and also better than the neural type-based models having no contextual information. In comparison to a sentence-based ByT5 Transformer fine-tuned through a text-translation fashion (Transnormer; \cite{bracke}), the performance of the smaller hybrid model is only on par, with the hybrid one slightly better in handling character replacements, but slightly worse in handling compound writing. In total, the system was unable to outperform the sentence-based one. 

However, the increasing accuracies of normalization systems also showed us the limits of the DTA EvalCorpus and the experimental setup. For one, we face inconsistencies in the corpus concerning normalization. For another, because taking the target side of corpus as the only “true” normalization form results in an overly restrictive conception of normalization, discarding other equally suitable normalization forms. Collectively, the findings regarding the systems' performances could be attributed to the shared performance limit ceiling by the dataset and experimental conditions, as both the Transnormer and the proposed hybrid system achieved similar results.
Having no large high-quality parallel corpus available remains a large obstacle to automatic historical text normalization.

Concerning the research question concerning type-based vs.\@ sentence-based models, we can interpret this evaluation in two complementing perspectives:
\begin{itemize}
    \item At least in the domain handled here, where word order is stable, combining type- and sentence-based models to a hybrid system \emph{does} help, requires little compute resources, but it requires sophisticated model designs. Actually, the easiest way to improve accuracy in text normalization is by starting from a large substitution lexicon for known words, extracted from parallel data.
    \item Smaller hybrid models are as accurate as larger sentence-based pre-trained Transformer models like ByT5, which have been fine-tuned to the text normalization task.
        However, hybrid models are much more sophisticated, whereas, e.g., ByT5, achieves considerate performance with minimal setup and large flexibility, since word order need not to be stable and tokenization, spacing, etc.\@ is handled end-to-end. But this comes with the drawback that these models require more decoding steps and more compute resources.
\end{itemize}
However, with the limited experiment presented here, a clear answer to the research question still remains inconclusive, with the largest limitation being dataset quality. As future work, a more thorough investigation could also incorporate the following variables:

First, how do the different system perform on different datasets and other languages? \textcite{bollmann_large-scale_2019}, for instance, tests systems on the normalization of historical English, German, Hungarian, Icelandic, Spanish, Portuguese and Swedish. The German datasets (Anselm Corpus, \cite{dipper_anselm_2013}, and RIDGES Herbology Corpus, \cites{odebrecht_ridges_2017}{ridges}) appear to be a very interesting case for first investigations: source text in Early New High German, but differences in register and normalization guidelines, and additionally the target text is \emph{not} contemporary standard German but a pseudo-German based on modern orthography but archaic syntax and extinct lexemes. But again, the availability of high-quality parallel corpora remains an issue.

Second, how dependent are the different system on the amount of training data they see? This would be particularly relevant for smaller projects of digital editions, who cannot resort to high-resource parallel data but must normalize a portion of their corpus by hand in order to obtain appropriate training data. Especially in a low-resource domain, I would suspect to see larger sentence-based models struggling to generalize in comparison to the much smaller type-based models. 

Third, what kind of variation do we see when modifying the model size and the training regime? For instance, what is the impact of the language model size, both in the proposed hybrid model and a sentence-based setup? Can the systems benefit from a larger language model? Or, going in the other direction, does the hybrid system maintain the same accuracy when going for a smaller (perhaps even $n$-gram-based) language model, coming with the benefit of faster inference? Concerning both the hybrid system and the sentence-based systems, experiments could also cover domain adaptation \parencites{gururangan_dont_2020,konle_domain_2020} of the language model, using (non-parallel) data from the target side, and in the case of the sentence-based Transnormer model even from the source side.

Finally, in the general context of historical text normalization, another point to be investigated would be the effect size of this normalization on downstream tasks. Does normalizing historical text actually improve the accuracy of NLP tools? As outlined in the introduction, we have evidence that text normalization is beneficial for basic NLP tasks such as Part-of-Speech Tagging, Named Entity Recognition, etc., however it remains open whether this also holds for tasks more present in CLS workflows, such as scene segmentation \parencite{zehe_detecting_2021,zehe_shared_2021} or the recognition of speech, thought, writing \parencite{brunner_bert_2021} in novels; tasks that particularly align with the literary domain considered here.
Corresponding experiments could also investigate whether cross-domain learning of pre-trained (multilingual) LLMs might be a viable alternative option, e.g. training on labeled texts in contemporary form and inference on texts in historic form.


\printbibliography[heading=bibintoc]\clearpage

\onecolumn
\appendix

\section{Dataset Split}\label{sec:split}
\bgroup
\begin{longtable}[l]{l@{\hspace*{3pt}}l@{}rS[table-format=7.0,table-text-alignment=right]r}
    \hline
\textbf{Author} & \textbf{Title} & \textbf{Year} & {\textbf{Tokens}} & \textbf{Split}\\
    \hline
    \endhead
\hline\endfoot
\bottomrule\endlastfoot
Iffland & Die Jäger & 1785 & 28064 & train \\
Schiller & Dom Karlos & 1787 & 37179 & train \\
Goethe & Iphigenie auf Tauris & 1787 & 14754 & train \\
Goethe & Torquato Tasso & 1790 & 25637 & train \\
Kotzebue & Menschenhaß und Reue & 1790 & 25267 & train \\
Klinger & Fausts Leben, Thaten und Höllenfahrt & 1791 & 62291 & train \\
Tieck & William Lovell. Erster Band & 1795 & 42789 & train \\
Goethe & Wilhelm Meisters Lehrjahre. Erster Band & 1795 & 48427 & train \\
Goethe & Wilhelm Meisters Lehrjahre. Zweyter Band & 1795 & 49639 & train \\
Goethe & Wilhelm Meisters Lehrjahre. Dritter Band & 1795 & 49977 & train \\
Tieck & William Lovell. Zweyter Band & 1796 & 57479 & train \\
Tieck & William Lovell. Dritter Band & 1796 & 62225 & train \\
Goethe & Wilhelm Meisters Lehrjahre. Vierter Band & 1796 & 68777 & train \\
Kotzebue & Die deutschen Kleinstädter & 1803 & 16627 & train \\
Kleist & Amphitryon & 1807 & 17640 & train \\
Goethe & Faust & 1808 & 29984 & train \\
Kleist & Das Käthchen von Heilbronn oder die Feuerprobe & 1810 & 23920 & train \\
Kleist & Der zerbrochne Krug & 1811 & 15608 & train \\
Grimm & Kinder- und Haus-Märchen & 1812 & 95262 & train \\
Hoffmann & Die Elixiere des Teufels & 1815 & 60172 & train \\
Hoffmann & Die Elixiere des Teufels. Zweiter Theil & 1816 & 57053 & train \\
Arnim & Der tolle Invalide auf dem Fort Ratonneau & 1818 & 8677 & train \\
Kleist & Die Schlacht bei Fehrbellin & 1822 & 14420 & train \\
Eichendorff & Aus dem Leben eines Taugenichts und das Marmorbild & 1826 & 33187 & train \\
Heine & Reisebilder. Erster Theil & 1826 & 16117 & train \\
Heine & Reisebilder. Zweiter Theil & 1827 & 28102 & train \\
Heine & Reisebilder. Dritter Theil & 1830 & 50744 & train \\
Grabbe & Napoleon oder Die hundert Tage & 1831 & 34193 & train \\
Heine & Reisebilder & 1831 & 39992 & train \\
Börne & Briefe aus Paris. Erster Theil & 1832 & 39330 & train \\
Börne & Briefe aus Paris. Zweiter Theil & 1832 & 37886 & train \\
Börne & Briefe aus Paris. Dritter Theil & 1833 & 38542 & train \\
Börne & Briefe aus Paris & 1833 & 45886 & train \\
Börne & Briefe aus Paris. Fünfter Theil & 1834 & 34981 & train \\
Börne & Briefe aus Paris. Sechster Theil & 1834 & 36979 & train \\
Brentano & Geschichte vom braven Kasperl und dem schönen Annerl & 1838 & 13327 & train \\
Arnim & Die Günderode. Erster Theil & 1840 & 80592 & train \\
Arnim & Die Günderode. Zweiter Theil & 1840 & 58144 & train \\
Stifter & Bunte Steine. Erster Band & 1853 & 57937 & train \\
Stifter & Bunte Steine. Zweiter Band & 1853 & 56265 & train \\
Keller & Der grüne Heinrich. Erster Band & 1854 & 65078 & train \\
Keller & Der grüne Heinrich. Zweiter Band & 1854 & 76342 & train \\
Keller & Der grüne Heinrich. Dritter Band & 1854 & 57821 & train \\
Keller & Der grüne Heinrich. Vierter Band & 1855 & 79084 & train \\
Ludwig & Zwischen Himmel und Erde & 1856 & 76225 & train \\
François & Die letzte Reckenburgerin. Erster Band & 1871 & 49814 & train \\
François & Die letzte Reckenburgerin. Zweiter Band & 1871 & 51017 & train \\
Spyri & Heidi's Lehr- und Wanderjahre & 1880 & 55268 & train \\
Fontane & Irrungen, Wirrungen & 1888 & 55451 & train \\
Raabe & Stopfkuchen & 1891 & 69149 & train \\
Fontane & Der Stechlin & 1899 & 137429 & train \\
Hofmannsthal & Tod des Tizian & 1901 & 2459 & train \\
    \hline
\textbf{Total} & & \emph{({\SI{59.61}{\percent}})} &2389022 & \textbf{train} \\
    \hline
    Heinse & Ardinghello und die glückseeligen Inseln. Erster Band & 1787 & 55924 & dev \\
Heinse & Ardinghello und die glückseeligen Inseln. Zweyter Band & 1787 & 51788 & dev \\
Tieck & Franz Sternbalds Wanderungen. Erster Theil & 1798 & 18934 & dev \\
Tieck & Franz Sternbald's Wanderungen. Zweiter Theil & 1798 & 45920 & dev \\
Fouqué & Undine & 1811 & 28034 & dev \\
Chamisso & Peter Schlemihl's wundersame Geschichte & 1814 & 21140 & dev \\
Hoffmann & Nachtstücke. Erster Theil & 1817 & 49127 & dev \\
Hoffmann & Nachtstücke. Zweiter Theil & 1817 & 64161 & dev \\
Arnim & Goethe's Briefwechsel mit einem Kinde. Erster Theil & 1835 & 59167 & dev \\
Arnim & Goethe's Briefwechsel mit einem Kinde. Zweiter Theil & 1835 & 58869 & dev \\
Arnim & Tagebuch. Tagebuch & 1835 & 44945 & dev \\
Stifter & Der Nachsommer. Erster Band & 1857 & 99839 & dev \\
Stifter & Der Nachsommer. Zweiter Band & 1857 & 88113 & dev \\
Stifter & Der Nachsommer. Dritter Band & 1857 & 94368 & dev \\
Keller & Das Sinngedicht & 1882 & 94356 & dev \\
Holz & Papa Hamlet & 1889 & 29525 & dev \\
Schnitzler & Liebelei & 1896 & 13193 & dev \\
    \hline
\textbf{Total} & & \emph{({\SI{22.89}{\percent}})} &917360 & \textbf{dev} \\
    \hline
    Schiller & Kabale und Liebe & 1784 & 30879 & test \\
Schiller & Der Geisterseher & 1789 & 39080 & test \\
Wackenroder & Herzensergießungen eines kunstliebenden Klosterbruders & 1797 & 32194 & test \\
Goethe & Die Wahlverwandtschaften. Erster Theil & 1809 & 42008 & test \\
Goethe & Die Wahlverwandtschaften. Zweyter Theil & 1809 & 46663 & test \\
Grimm & Kinder- und Haus-Märchen. Zweiter Band & 1815 & 74176 & test \\
Hoffmann & Meister Floh & 1822 & 43319 & test \\
Hauff & Phantasien im Bremer Rathskeller & 1827 & 18707 & test \\
Grillparzer & Ein treuer Diener seines Herrn & 1830 & 10951 & test \\
Gutzkow & Wally, die Zweiflerin & 1835 & 39853 & test \\
Hebbel & Maria Magdalene & 1844 & 15638 & test \\
Storm & Immensee & 1852 & 11024 & test \\
Keller & Die Leute von Seldwyla & 1856 & 78447 & test \\
Raabe & Die Chronik der Sperlingsgasse & 1857 & 50597 & test \\
Raabe & Das Odfeld & 1889 & 60196 & test \\
Fontane & Effi Briest & 1896 & 107611 & test \\
    \hline
\textbf{Total} & & \emph{({\SI{17.50}{\percent}})} &701318 & \textbf{test} \\
    \hline
\textbf{Total} & & &4007700& \\
\end{longtable}
\egroup

\end{document}